\documentclass[letterpaper,10pt,journal,twoside]{IEEEtran}

\usepackage{amsmath,amssymb}
\usepackage{graphicx}
\usepackage{booktabs}
\usepackage{multirow}
\usepackage{array}
\usepackage{xcolor}
\usepackage{cite}
\usepackage{url}
\usepackage{xspace}
\usepackage[caption=false,font=footnotesize]{subfig}
\usepackage{algorithm}
\usepackage{algorithmic}

\graphicspath{{figures/}}

\newcommand{\method}{HCPG-Flow\xspace}

\newcommand{\metaworld}{MetaWorld\xspace}
\newcommand{\maniskill}{ManiSkill\xspace}
\newif\ifrealresults
\realresultstrue
\newif\ifprojectpagepublic
\projectpagepublicfalse
\newcommand{\projectpageurl}{https://hitxraz.github.io/HCPG-Flow/}

\title{\method: Hierarchical Contact-Progress Guidance for Flow-Policy Robot Manipulation}

\author{Guanghu Xie, Mingxu Li, Shuo Zhang, Yonglong Zhang, Yifan Yang,
Yang Liu$^{*}$, Zongwu Xie, Baoshi Cao%
\thanks{This work was supported by the Natural Science Foundation of Heilongjiang Province for Excellent Young Scholars (Grant No. YQ2024E018) and the Youth Talent Support Program of China (Grant No. 2022-JCJQ-QT-061).}%
\thanks{The authors are with the State Key Laboratory of Robotics and Systems, Harbin Institute of Technology, Harbin 150001, Heilongjiang, China (e-mail: 23b308003@stu.hit.edu.cn; 25S008014@stu.hit.edu.cn; 25S008055@stu.hit.edu.cn; zhangyl202601@163.com; yangyifan03001@163.com; liuyanghit@hit.edu.cn; xiezongwu@hit.edu.cn; cbs@hit.edu.cn).}%
\thanks{$^{*}$Corresponding author: Yang Liu (e-mail: liuyanghit@hit.edu.cn).}}

\begin{document}

\maketitle

\ifprojectpagepublic
\begin{center}
  \footnotesize Project page: \url{\projectpageurl}
\end{center}
\vspace{-0.6em}
\fi

\begin{abstract}
Flow policies can represent multimodal action distributions for robot manipulation, yet a robot must execute one action at each control step.
When several proposals are sampled, critic-based ranking makes data collection depend on value estimates over candidate actions that may be weakly represented in replay.
We introduce \method, an analytic rollout-time selector that augments SAC-Flow with hierarchical, object-centric contact-progress guidance while preserving its actor and critic objectives.
HCPG switches from end-effector approach to task progress after contact, scores each proposal by the first-order reduction of a task-relevant distance, standardizes scores within the candidate set, and executes a temperature-controlled action embedding.
Across ten simulated tasks, HCPG improves mean success over SAC-Flow on both benchmarks, including a 9.5 percentage-point gain on \maniskill.
Four physical tasks further show high success with a 17.4\% reduction in successful completion steps.
\end{abstract}

\begin{IEEEkeywords}
Reinforcement learning, robot manipulation, flow policies.
\end{IEEEkeywords}

\section{Introduction}
\IEEEPARstart{R}{obot} manipulation couples multimodal action choices with abrupt changes in contact.
Flow- and diffusion-based policies are well suited to this setting because they can represent action distributions that a unimodal Gaussian policy may not capture~\cite{ho2020ddpm,lipman2023flowmatching,liu2023rectifiedflow,chi2023diffusionpolicy}.
SAC-Flow~\cite{zhang2025sacflow} combines this expressive policy class with Soft Actor-Critic (SAC)~\cite{haarnoja2018sac}, allowing several plausible actions to be generated from the same state.
This capability shifts part of the control problem from generating an action to deciding which proposal should be executed.

Sampling several actions from a generative policy and selecting the highest-valued proposal is an established strategy in batch-constrained and diffusion-based RL~\cite{fujimoto2019bcq,mao2024diffusiondice}.
This rule reuses the learned value function, but its decision quality inherits critic errors, particularly for actions weakly supported by the training distribution~\cite{fujimoto2019bcq,wang2023diffusionql,mao2024diffusiondice}.
This issue is especially relevant around grasp, push, insertion, and release transitions, where similar long-horizon estimates can correspond to different immediate geometric consequences.
Our experiments likewise show that Q-guided candidate selection is not consistently stronger than single-candidate SAC-Flow.
The unresolved question is therefore how to exploit proposal diversity using information that is available before reliable long-horizon ranking.

Our key observation is that many manipulation tasks expose a local progress signal through the geometry among the tool center point (TCP), manipulated object, and task goal~\cite{zeng2021transporter,chi2023diffusionpolicy,chi2024umi}.
Before contact, useful actions move the TCP toward the object; after contact, they move the object along a task-relevant direction.
\method turns this observation into a hierarchical candidate selector.
It uses a contact-aware phase gate to switch between approach and task progress, evaluates each proposal through a first-order distance-reduction score, calibrates scores within the proposal set, and forms a soft action embedding.
The resulting module operates only during rollout and leaves the actor and critic objectives unchanged.

The principal contributions are:
\begin{itemize}
  \item \textbf{An object-centric formulation of flow-policy candidate selection.}
  We formulate execution-time selection as phase-conditioned object-centric progress and relate the directional score to the local first-order reduction of a task distance.
  \item \textbf{An analytic selector that preserves the learning backbone.}
  HCPG combines a contact-aware phase gate, within-set score normalization, and soft action embedding without introducing a learned scorer, auxiliary loss, or additional backward pass.
  \item \textbf{Empirical gains across simulation and real-world manipulation.}
Across ten simulated tasks, HCPG improves the task-averaged success rate of SAC-Flow on both ManiSkill and MetaWorld, including a 9.5 percentage-point gain on ManiSkill. Across four physical tasks, it raises aggregate success from 91.7\% to 98.3\% and reduces successful completion steps by 17.4\%.
\end{itemize}

\begin{figure*}[t]
  \centering
  \includegraphics[width=\textwidth]{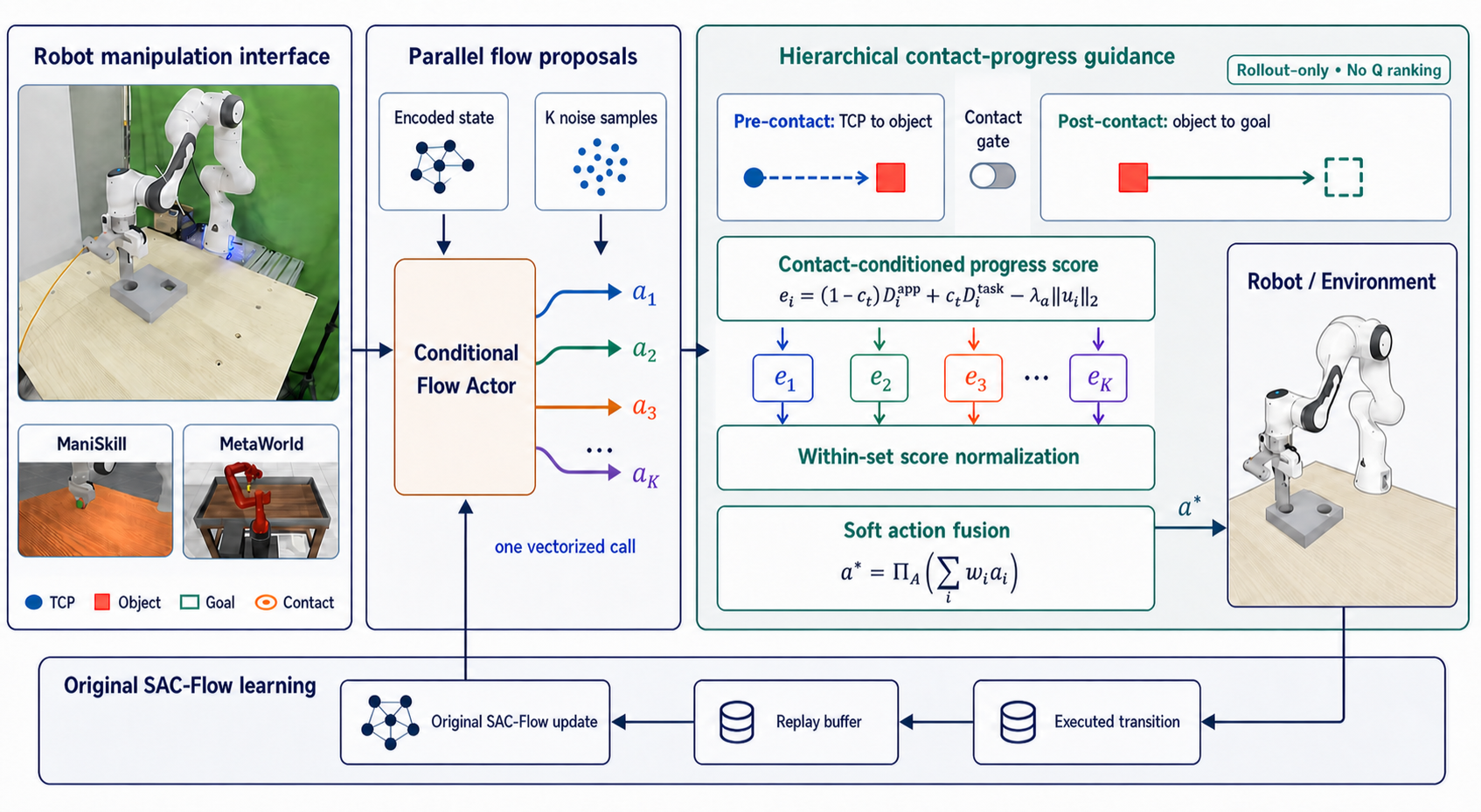}
  \caption{Robot-centric overview of \method. Physical and simulated manipulation tasks provide state and contact cues to a conditional flow actor, which generates $K=4$ full-dimensional action proposals in one vectorized call. HCPG switches the active progress target at contact, scores the proposals, and forms a bounded soft action without critic-based ranking. The executed transition enters the original replay buffer and SAC-Flow update; the learning objectives remain unchanged.}
  \label{fig:framework}
\end{figure*}

\section{Related Work}

\subsection{Maximum-Entropy RL and Generative Policies}
Policy-gradient and actor-critic methods provide the optimization backbone for continuous-control RL~\cite{sutton1999policygradient,lillicrap2016ddpg,fujimoto2018td3,schulman2017ppo}.
SAC~\cite{haarnoja2018sac} combines off-policy value learning with entropy-regularized policy improvement, but its standard Gaussian actor can be restrictive for multimodal action distributions.
Diffusion and flow models provide more expressive alternatives for decision making and robot control~\cite{ho2020ddpm,lipman2023flowmatching,liu2023rectifiedflow,florence2022ibc,janner2022diffuser,ajay2023decisiondiffuser,wang2023diffusionql,chi2023diffusionpolicy,ren2024dppo}.
SAC-Flow~\cite{zhang2025sacflow}, FlowRL~\cite{lv2025flowrl}, QSM~\cite{psenka2024qsm}, and DIME~\cite{celik2025dime} study flow- or score-based policy parameterizations and their optimization.

\subsection{Candidate Evaluation and Goal-Conditioned Control}
Value-based candidate ranking selects actions according to predicted long-horizon return, as used broadly in actor-critic and diffusion-based RL~\cite{haarnoja2018sac,wang2023diffusionql,psenka2024qsm}.
Model-based planning provides another route by simulating candidate consequences, but it requires a dynamics model and additional rollout computation.
Hindsight Experience Replay~\cite{andrychowicz2017her} and related goal-conditioned methods use achieved goals to improve learning under sparse rewards.

\subsection{Object-Centric Robot Manipulation}
Relative object geometry is a recurring inductive bias in visual manipulation and large-scale robot learning~\cite{zeng2021transporter,brohan2023rt1,zhao2023aloha,chi2024umi}.
The same end-effector displacement can be useful or harmful depending on the contact phase and object-goal relation, motivating action evaluation in object-centric coordinates.

\section{Method}

\subsection{SAC-Flow Backbone and Problem Formulation}
We consider a continuous-control Markov decision process
$\mathcal{M}=(\mathcal{S},\mathcal{A},P,r,\gamma)$ with replay buffer
$\mathcal{D}$.
SAC-Flow replaces the Gaussian SAC actor with a conditional flow policy while retaining off-policy twin-critic learning and entropy-regularized policy improvement~\cite{zhang2025sacflow}.
Starting from base noise $x_i^0=\epsilon_i\sim p_0$, its velocity-reparameterized actor generates an action through $M$ integration steps,
\begin{equation}
  x_i^{m+1}=x_i^m+\Delta t\,v_\theta(x_i^m,t_m;s_t),
  \qquad
  a_t^i=\Pi_{\mathcal A}(x_i^M),
  \label{eq:sacflow-sampling}
\end{equation}
where $v_\theta$ is the conditional velocity network and $\Pi_{\mathcal A}$ enforces the action bounds.
The differentiable flow rollout receives the SAC actor signal from the learned critics; executed transitions are stored in $\mathcal D$, and the actor and critics are updated using the original SAC-Flow objectives.

For HCPG-Flow, the same actor maps independent noise samples to a candidate set,
\begin{equation}
  a_t^{i}=F_\theta(\epsilon_i;s_t),\qquad
  \epsilon_i\sim p_0,\quad i\in\{1,\ldots,K\},
  \label{eq:flow-candidates}
\end{equation}
where $F_\theta$ denotes the complete integration map in Eq.~\eqref{eq:sacflow-sampling}.
Sampling is vectorized by repeating the state features $K$ times in one actor call, yielding
$\mathcal{A}_t^K=\{a_t^1,\ldots,a_t^K\}$.
The SAC-Flow baseline executes its single sample directly.

A critic-guided selector would rank this set using
$\widehat Q_\phi(s_t,a_t^i)$, making the collected transition depend on the critic's ordering over the proposals.
We instead seek a rollout-time map
$G(s_t,\mathcal{A}_t^K)\mapsto a_t^\star$
that uses task geometry available in the observation, adds no learned scorer, remains comparable across candidates and task scales, and leaves the SAC-Flow losses unchanged.
As shown in Fig.~\ref{fig:framework}, HCPG implements this map in three steps: it constructs object-centric approach and task directions, evaluates candidates with a contact-conditioned progress score, and converts within-set normalized scores into a bounded action.

\subsection{Object-Centric Task Adapter}
The task adapter exposes the local geometry required by the selector.
It maps the raw observation into
\begin{equation}
  \phi(s_t)=
  \left(x_t^{\mathrm{tcp}},x_t^o,x_t^g,z_t\right),
  \label{eq:task-adapter}
\end{equation}
where $x_t^{\mathrm{tcp}}$ is the end-effector or tool position,
$x_t^o$ is the manipulated-object position,
$x_t^g$ is the goal or task reference, and $z_t$ contains an optional grasp/contact indicator.
Only the translational part of each candidate,
$u_i=\operatorname{xyz}(a_t^i)$, is used to compute the progress score.
Rotational and gripper components remain part of the flow-generated candidate and are retained by the soft action embedding described below.

We define an approach vector and a task vector,
\begin{equation}
  \rho_t^{\mathrm{app}}=x_t^o-x_t^{\mathrm{tcp}},
  \qquad
  \rho_t^{\mathrm{task}}=x_t^g-x_t^o.
  \label{eq:relative-vectors}
\end{equation}
The adapter specifies the semantic interpretation of these vectors for each task family rather than changing the underlying state seen by SAC-Flow.
For example, PokeCube uses the peg head as the tool point, while lifting uses the vertical axis as the post-contact task direction.


\subsection{First-Order Directional Progress}
For a candidate translation $u$ and target vector $\rho$, define
\begin{equation}
  D(u,\rho)=
  \frac{u^\top\rho}{\max(\|\rho\|_2,\varepsilon)}.
  \label{eq:directional-progress}
\end{equation}
The directional score is a local surrogate for immediate geometric progress.
Let $d(x,g)=\|g-x\|_2$ and assume a local displacement
$x^+=x+\alpha u$ with small $\alpha>0$.
A first-order expansion gives
\begin{equation}
  d(x,g)-d(x+\alpha u,g)
  =\alpha D(u,g-x)+\mathcal{O}(\alpha^2).
  \label{eq:first-order}
\end{equation}
Thus, under this local displacement model, maximizing $D$ maximizes the first-order reduction in target distance.
Before contact, this argument applies to TCP-object distance.
After contact, it applies to object-goal distance under the local approximation that TCP displacement transfers to the object.
This interpretation motivates candidate ranking but does not assume that distance reduction is globally optimal over a long horizon.

\subsection{Hierarchical Contact-Progress Score}
The geometric objective changes when the robot establishes contact or grasps the object.
HCPG represents this two-level hierarchy using a phase variable
\begin{equation}
  c_t=
  \begin{cases}
    \operatorname{clip}(z_t^{\mathrm{grasp}},0,1),
      & \text{if a grasp flag is available},\\
    \mathbb{I}\!\left[\|\rho_t^{\mathrm{app}}\|_2<d_c\right],
      & \text{otherwise}.
  \end{cases}
  \label{eq:phase-gate}
\end{equation}
The candidate progress score is
\begin{equation}
  e_i=
  (1-c_t)D(u_i,\rho_t^{\mathrm{app}})
  +c_tD(u_i,\rho_t^{\mathrm{task}})
  -\lambda_a\|u_i\|_2.
  \label{eq:hcpg-score}
\end{equation}
The first term rewards approach before contact, the second rewards task progress after contact, and the final term breaks otherwise similar rankings in favor of smaller translational commands.
The main experiments use $d_c=0.08$ and $\lambda_a=0.05$ for all tasks.
The lift adapter replaces $\rho_t^{\mathrm{task}}$ with the positive vertical axis, while articulated tasks use the task axis or goal displacement supplied by the observation adapter.

\subsection{Within-Set Score Normalization and Soft Embedding}
Raw progress scores vary with task scale and the local state.
HCPG therefore standardizes them within each proposal set:
\begin{equation}
  \widehat e_i=
  \frac{e_i-\mu_e}{\max(\sigma_e,\varepsilon)},\quad
  \mu_e=\frac{1}{K}\sum_{j=1}^{K}e_j,
  \label{eq:candidate-normalization}
\end{equation}
where $\sigma_e$ is the corresponding population standard deviation.
This normalization preserves the within-set ordering while making the temperature less sensitive to the absolute score scale.

Rather than executing a hard $\arg\max$, HCPG forms a temperature-controlled action embedding,
\begin{equation}
  \omega_i=
  \frac{\exp(\widehat e_i/\tau)}
       {\sum_{j=1}^{K}\exp(\widehat e_j/\tau)},\qquad
  a_t^\star=
  \Pi_{\mathcal{A}}\left(\sum_{i=1}^{K}\omega_i a_t^i\right),
  \label{eq:soft-embedding}
\end{equation}
where $\Pi_{\mathcal A}$ clips the blended action to the valid action bounds.
This mapping is continuous with respect to the candidate scores and keeps the executed action in the convex hull of the proposals before clipping.
We use $K=4$, $\tau=0.7$, and no top-$k$ truncation in the main experiments.

\begin{algorithm}[t]
  \caption{\method rollout-time candidate selection}
  \label{alg:hcpg}
  \begin{algorithmic}[1]
    \REQUIRE state $s_t$, flow actor $F_\theta$, adapter $\phi$, $K$, $d_c$, $\lambda_a$, $\tau$
    \STATE Sample $a_t^{1:K}=F_\theta(\epsilon_{1:K};s_t)$ in one vectorized call
    \STATE Extract $\rho_t^{\mathrm{app}}$, $\rho_t^{\mathrm{task}}$, and phase $c_t$ using Eqs.~\eqref{eq:relative-vectors}--\eqref{eq:phase-gate}
    \FOR{$i=1,\ldots,K$}
      \STATE Compute progress score $e_i$ using Eq.~\eqref{eq:hcpg-score}
    \ENDFOR
    \STATE Standardize $\{e_i\}$ using Eq.~\eqref{eq:candidate-normalization}
    \STATE Compute weights $\omega_{1:K}$ and action $a_t^\star$ using Eq.~\eqref{eq:soft-embedding}
    \STATE Execute $a_t^\star$ and store $(s_t,a_t^\star,r_t,s_{t+1})$ in $\mathcal{D}$
    \STATE Update the SAC critic and SAC-Flow actor without modification
  \end{algorithmic}
\end{algorithm}

\section{Experiments}

\subsection{Evaluation Protocol}
We evaluate on four \maniskill tasks: PickCube, PushCube, PokeCube, and PullCube.
These tasks cover grasping, pushing, poking, and pulling with explicit object-goal structure.
We also evaluate on six \metaworld tasks: ButtonPress, DrawerOpen, DoorOpen, SweepInto, PegInsertSide, and LeverPull.
These tasks include contact-rich articulated manipulation and object relocation.


Our primary baseline is SAC-Flow, which retains the same flow-policy learning backbone but does not use multi-candidate HCPG selection~\cite{zhang2025sacflow}.
The \metaworld comparison additionally includes SAC~\cite{haarnoja2018sac}, FlowRL~\cite{lv2025flowrl}, and QSM~\cite{psenka2024qsm}; the \maniskill table includes FlowRL. 
We report mean and standard deviation over protocol-matched seeds 0, 1, and 2 using the final five evaluations.

\subsection{Implementation Details}
Simulation experiments run on NVIDIA GeForce RTX 4090 GPUs.
HCPG uses $K=4$, $\tau=0.7$, $d_c=0.08$, and $\lambda_a=0.05$.
Within each task, all methods use the same training budget, environment parallelism, update schedule, and evaluation frequency.
Real-robot policies issue Cartesian end-effector delta commands at 30 Hz under matched sensing and control settings.

\subsection{Main Results}
Across both suites, HCPG improves the across-task mean of its matched SAC-Flow backbone and remains within 0.1 percentage points on saturated tasks.
On \metaworld (Table~\ref{tab:metaworld-main}), HCPG raises the across-task mean from 94.7\% to 97.1\%.
It improves SAC-Flow on ButtonPress, SweepInto, PegInsertSide, and LeverPull, with gains of 0.9--9.3 percentage points, and remains within 0.1 points on the saturated DrawerOpen and DoorOpen tasks.
HCPG has the highest across-task mean among the compared methods, although QSM is strongest on ButtonPress and PegInsertSide.

On \maniskill (Table~\ref{tab:maniskill-main}), HCPG raises the across-task mean from 87.2\% to 96.7\%.
The largest gain is 34.2 points on PickCube, where the selector must distinguish approach, grasp, and transport proposals.
The remaining three tasks already exceed 96\% under SAC-Flow; HCPG adds 0.4--2.1 points.
FlowRL performs strongly on these saturated tasks but fails to learn PickCube reliably, yielding a lower across-task mean.

\begin{table*}[t]
  \centering
\caption{MetaWorld last-five episode success (\%). Values are reported as mean $\pm$ SD. Task rows use SD over seeds 0/1/2; the final row uses sample SD across task means. SD quantifies variability and is not a confidence interval or a bounded success-rate range.}
  \label{tab:metaworld-main}
  \resizebox{\textwidth}{!}{%
  \begin{tabular}{lccccc}
    \toprule
    Task & SAC & FlowRL & QSM & SAC-Flow & HCPG-Flow \\
    \midrule
    ButtonPress & 82.0 $\pm$ 13.2 & 78.0 $\pm$ 8.4 & \textbf{94.5 $\pm$ 5.2} & 82.5 $\pm$ 8.2 & 91.9 $\pm$ 2.3 \\
    DrawerOpen & 55.1 $\pm$ 48.9 & 65.3 $\pm$ 56.6 & 33.3 $\pm$ 57.7 & \textbf{100.0 $\pm$ 0.0} & 99.9 $\pm$ 0.2 \\
    DoorOpen & 99.9 $\pm$ 0.2 & 99.5 $\pm$ 0.6 & \textbf{100.0 $\pm$ 0.0} & \textbf{100.0 $\pm$ 0.0} & 99.9 $\pm$ 0.2 \\
    SweepInto & 83.9 $\pm$ 8.5 & 83.1 $\pm$ 14.0 & 51.7 $\pm$ 46.0 & 92.3 $\pm$ 3.6 & \textbf{93.7 $\pm$ 6.7} \\
    PegInsertSide & 79.7 $\pm$ 31.7 & 64.8 $\pm$ 56.1 & \textbf{99.9 $\pm$ 0.2} & 98.1 $\pm$ 2.6 & 99.1 $\pm$ 1.0 \\
    LeverPull & 55.9 $\pm$ 46.7 & 93.2 $\pm$ 8.4 & 64.9 $\pm$ 38.9 & 95.1 $\pm$ 1.8 & \textbf{98.4 $\pm$ 2.4} \\
    \midrule
    Across-task mean & 76.1 $\pm$ 17.5 & 80.6 $\pm$ 14.2 & 74.1 $\pm$ 28.3 & 94.7 $\pm$ 6.7 & \textbf{97.1 $\pm$ 3.5} \\
    \bottomrule
  \end{tabular}}
\end{table*}

\begin{table}[t]
  \centering
  \caption{ManiSkill last-five episode success (\%). Values are reported as mean $\pm$ SD. Task rows use SD over seeds 0/1/2; the final row uses sample SD across task means. SD quantifies variability and is not a confidence interval or a bounded success-rate range. QSM is omitted because its critic-gradient target was numerically unstable under this protocol.}
  \label{tab:maniskill-main}
  \resizebox{\columnwidth}{!}{%
  \begin{tabular}{lccc}
    \toprule
    Task & FlowRL & SAC-Flow & HCPG-Flow \\
    \midrule
    PickCube & 1.6 $\pm$ 1.4 & 57.5 $\pm$ 14.2 & \textbf{91.7 $\pm$ 8.8} \\
    PushCube & \textbf{99.9 $\pm$ 0.2} & 96.7 $\pm$ 2.9 & 98.8 $\pm$ 1.3 \\
    PokeCube & \textbf{96.8 $\pm$ 1.6} & 96.2 $\pm$ 3.3 & 96.7 $\pm$ 1.4 \\
    PullCube & \textbf{99.9 $\pm$ 0.2} & 98.3 $\pm$ 1.9 & 99.6 $\pm$ 0.7 \\
    \midrule
    Across-task mean & 74.5 $\pm$ 48.6 & 87.2 $\pm$ 19.8 & \textbf{96.7 $\pm$ 3.6} \\
    \bottomrule
  \end{tabular}}
\end{table}

\begin{figure*}[t]
  \centering
  \includegraphics[width=\textwidth]{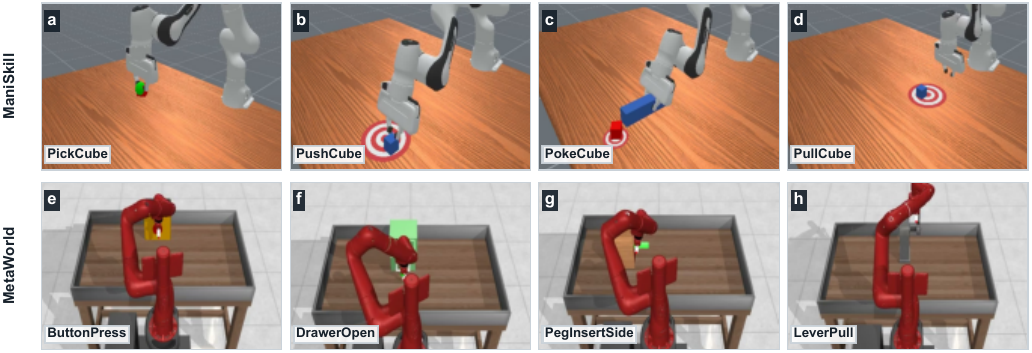}
  \caption{Representative successful simulation rollouts. The top row shows \maniskill manipulation tasks; the bottom row shows \metaworld contact-rich and articulated manipulation tasks.}
  \label{fig:simulation-rollouts}
\end{figure*}

\begin{figure}[t]
  \centering
  \includegraphics[width=\columnwidth]{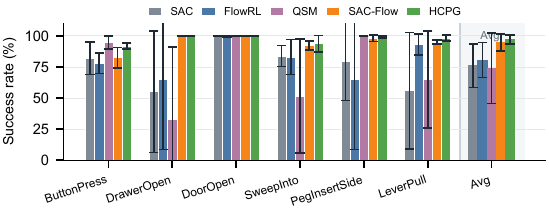}
  \caption{MetaWorld last-five episode success. Bars show means and black whiskers show $\pm$1 SD as variability, not confidence intervals or bounded success-rate ranges; the shaded Avg group reports the across-task mean.}
  \label{fig:metaworld-bars}
\end{figure}

\begin{figure}[t]
  \centering
  \includegraphics[width=\columnwidth]{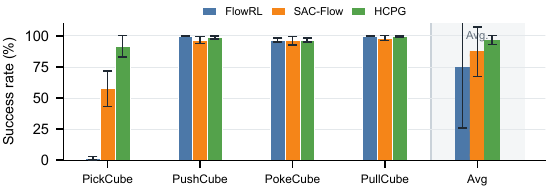}
  \caption{ManiSkill last-five episode success. Bars show means and black whiskers show $\pm$1 SD as variability, not confidence intervals or bounded success-rate ranges; the shaded Avg group reports the across-task mean.}
  \label{fig:maniskill-bars}
\end{figure}

\begin{figure}[t]
  \centering
  \includegraphics[width=\columnwidth]{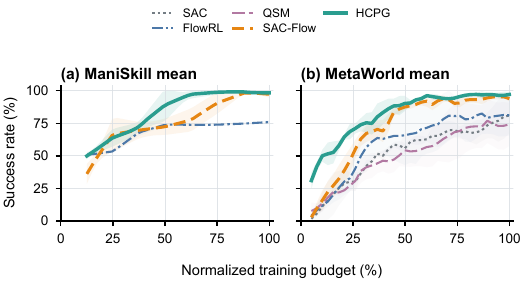}
  \caption{Simulation learning progress for all main-table methods. Each panel averages all tasks in the corresponding suite within each seed after linearly aligning normalized training budgets. Lines show means over seeds 0, 1, and 2; shaded regions denote $\pm$1 SD.}
  \label{fig:sample-efficiency}
\end{figure}

Figure~\ref{fig:sample-efficiency} reports checkpoint-level evaluation success for both simulation suites under the same fixed protocol as the main tables.
HCPG improves the average learning profile across most of the training budget.

\subsection{Ablation Study}
\label{sec:ablation}
We isolate the selection rule and candidate count on PickCube, PushCube, and PullCube.
All variants have complete coverage over three tasks and three seeds.
As shown in Table~\ref{tab:ablation-decision}, HCPG K=4 achieves 96.7\% last-five success and 85.2 success AUC, compared with 84.2\% and 75.1 for SAC-Flow K=1.
Q-guided K=4 attains high final success but lower last-five success (81.8\%) and AUC (71.4), showing that four proposals and critic ranking alone do not reproduce the HCPG result.
Table~\ref{tab:k-sensitivity} further shows that increasing $K$ is not sufficient: K=8 does not dominate K=4, whereas K=2 and K=8 are markedly less reliable on PickCube.
We therefore use K=4 throughout the main comparison.

\begin{table}[t]
  \centering
  \caption{Selection and candidate-count ablation on PickCube, PushCube, and PullCube over seeds 0/1/2. Metrics are averaged across the nine task-seed runs.}
  \label{tab:ablation-decision}
  \resizebox{\columnwidth}{!}{%
  \begin{tabular}{lcccc}
    \toprule
    Method & Runs & Last-5 & AUC & Final \\
    \midrule
    SAC-Flow & 9/9 & 84.2 & 75.1 & 96.5 \\
    Q-guided K=4 & 9/9 & 81.8 & 71.4 & \textbf{99.3} \\
    HCPG K=2 & 9/9 & 80.8 & 73.8 & 98.6 \\
    \textbf{HCPG K=4} & 9/9 & \textbf{96.7} & \textbf{85.2} & \textbf{99.3} \\
    HCPG K=8 & 9/9 & 80.8 & 74.1 & \textbf{99.3} \\
    \bottomrule
  \end{tabular}}
\end{table}

\begin{table}[t]
  \centering
  \caption{Sensitivity to candidate count $K$, measured by last-five success (\%). K=4 provides the strongest aggregate last-five success and AUC in Table~\ref{tab:ablation-decision}.}
  \label{tab:k-sensitivity}
  \resizebox{\columnwidth}{!}{%
  \begin{tabular}{lccc}
    \toprule
    Task & HCPG K=2 & HCPG K=4 & HCPG K=8 \\
    \midrule
    PickCube & 42.5 $\pm$ 13.5 & \textbf{91.7 $\pm$ 8.8} & 46.2 $\pm$ 15.7 \\
    PushCube & \textbf{100.0 $\pm$ 0.0} & 98.8 $\pm$ 1.3 & 98.8 $\pm$ 1.3 \\
    PullCube & \textbf{100.0 $\pm$ 0.0} & 99.6 $\pm$ 0.7 & 97.5 $\pm$ 2.2 \\
    \bottomrule
  \end{tabular}}
\end{table}

\ifrealresults
\section{Real-Robot Evaluation}
\label{sec:real}
We study peg insertion, button pressing, drawer closing, and rugby-ball grasping.
The tasks instantiate four contact-progress structures used by HCPG: constrained insertion, short-horizon pressing, articulated closing, and an approach--grasp transition.
Policies are initialized and trained under matched task budgets, then evaluated for 15 autonomous trials without intervention.
The budgets are 1200 environment steps for insertion, 1000 for button pressing and rugby grasping, and 400 for drawer closing.

\begin{table}[t]
  \centering
  \caption{Real-robot evaluation over 15 trials per task. Completion steps are averaged over successful trials and count actions to the first task-condition hit. Lower is preferred.}
  \label{tab:real-results}
  \resizebox{\columnwidth}{!}{%
  \begin{tabular}{lrrrr}
    \toprule
    & \multicolumn{2}{c}{Success (\%)} & \multicolumn{2}{c}{Completion steps} \\
    \cmidrule(lr){2-3}\cmidrule(lr){4-5}
    Task & SAC-Flow & HCPG-Flow & SAC-Flow & HCPG-Flow \\
    \midrule
    Peg insertion & 100.0 & 100.0 & 29.6 & 24.3 \\
    Button press & 100.0 & 100.0 & 32.1 & 31.3 \\
    Drawer closing & 100.0 & 100.0 & 127.7 & 107.4 \\
    Rugby grasp & 66.7 & 93.3 & 60.2 & 43.4 \\
    \midrule
    Overall / macro mean & 91.7 & \textbf{98.3} & 62.4 & \textbf{51.6} \\
    \bottomrule
  \end{tabular}}
\end{table}

Both methods solve all 15 insertion, button, and drawer trials.
HCPG reduces the mean steps to first success from 29.6 to 24.3 for insertion, from 32.1 to 31.3 for button pressing, and from 127.7 to 107.4 for drawer closing.
The largest difference appears on rugby grasping, where HCPG reaches 14/15 successes compared with 10/15 for SAC-Flow; among successful trials, completion steps decrease from 60.2 to 43.4.

Pooling the four tasks, HCPG records 59/60 successes (98.3\%) and SAC-Flow records 55/60 (91.7\%).
Across task-level means, successful completion steps decrease from 62.4 to 51.6 (17.4\%).
Figure~\ref{fig:real-rollouts} shows representative task-completion states from successful HCPG rollouts.

\begin{figure}[t]
  \centering
  \includegraphics[width=\columnwidth]{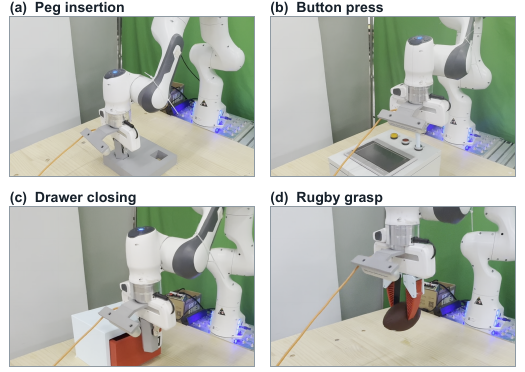}
  \caption{Successful HCPG-Flow completion states on the physical Franka platform: (a) the peg is fully inserted, (b) the button is depressed, (c) the drawer is closed, and (d) the rugby ball is securely grasped above the support surface.}
  \label{fig:real-rollouts}
\end{figure}
\fi

\section{Conclusion}
We introduced \method, an analytic candidate selector that converts flow-policy proposals into robot actions using contact-conditioned object-centric progress.
Across ten \maniskill and \metaworld tasks, HCPG improves the across-task success of its matched SAC-Flow backbone, and ablations distinguish the gain from candidate count and Q-guided ranking alone.
Four physical tasks further show improved pooled success and a 17.4\% reduction in successful completion steps.
Because the selector leaves the actor and critic objectives unchanged and adds no learned parameters, it provides a lightweight route for injecting manipulation geometry into online data collection.

\bibliographystyle{IEEEtran}
\bibliography{refs}

@inproceedings{sutton1999policygradient,
  author={Sutton, Richard S. and McAllester, David and Singh, Satinder and Mansour, Yishay},
  title={Policy Gradient Methods for Reinforcement Learning with Function Approximation},
  booktitle={Advances in Neural Information Processing Systems},
  year={1999}
}

@inproceedings{lillicrap2016ddpg,
  author={Lillicrap, Timothy P. and Hunt, Jonathan J. and Pritzel, Alexander and Heess, Nicolas and Erez, Tom and Tassa, Yuval and Silver, David and Wierstra, Daan},
  title={Continuous Control with Deep Reinforcement Learning},
  booktitle={International Conference on Learning Representations},
  year={2016}
}

@article{schulman2017ppo,
  author={Schulman, John and Wolski, Filip and Dhariwal, Prafulla and Radford, Alec and Klimov, Oleg},
  title={Proximal Policy Optimization Algorithms},
  journal={arXiv preprint arXiv:1707.06347},
  year={2017}
}

@inproceedings{andrychowicz2017her,
  author={Andrychowicz, Marcin and Wolski, Filip and Ray, Alex and Schneider, Jonas and Fong, Rachel and Welinder, Peter and McGrew, Bob and Tobin, Josh and Abbeel, Pieter and Zaremba, Wojciech},
  title={Hindsight Experience Replay},
  booktitle={Advances in Neural Information Processing Systems},
  year={2017}
}

@inproceedings{haarnoja2018sac,
  author={Haarnoja, Tuomas and Zhou, Aurick and Abbeel, Pieter and Levine, Sergey},
  title={Soft Actor-Critic: Off-Policy Maximum Entropy Deep Reinforcement Learning with a Stochastic Actor},
  booktitle={International Conference on Machine Learning},
  year={2018}
}

@inproceedings{fujimoto2018td3,
  author={Fujimoto, Scott and van Hoof, Herke and Meger, David},
  title={Addressing Function Approximation Error in Actor-Critic Methods},
  booktitle={International Conference on Machine Learning},
  year={2018}
}

@inproceedings{fujimoto2019bcq,
  author={Fujimoto, Scott and Meger, David and Precup, Doina},
  title={Off-Policy Deep Reinforcement Learning without Exploration},
  booktitle={Proceedings of the 36th International Conference on Machine Learning},
  volume={97},
  series={Proceedings of Machine Learning Research},
  pages={2052--2062},
  publisher={PMLR},
  year={2019}
}

@inproceedings{ho2020ddpm,
  author={Ho, Jonathan and Jain, Ajay and Abbeel, Pieter},
  title={Denoising Diffusion Probabilistic Models},
  booktitle={Advances in Neural Information Processing Systems},
  year={2020}
}

@inproceedings{lipman2023flowmatching,
  author={Lipman, Yaron and Chen, Ricky T. Q. and Ben-Hamu, Heli and Nickel, Maximilian and Le, Matthew},
  title={Flow Matching for Generative Modeling},
  booktitle={International Conference on Learning Representations},
  year={2023}
}

@inproceedings{liu2023rectifiedflow,
  author={Liu, Xingchao and Gong, Chengyue and Liu, Qiang},
  title={Flow Straight and Fast: Learning to Generate and Transfer Data with Rectified Flow},
  booktitle={International Conference on Learning Representations},
  year={2023}
}

@inproceedings{florence2022ibc,
  author={Florence, Pete and Lynch, Corey and Zeng, Andy and Ramirez, Oscar A. and Wahid, Ayzaan and Downs, Laura and Wong, Adrian and Lee, Johnny and Mordatch, Igor and Tompson, Jonathan},
  title={Implicit Behavioral Cloning},
  booktitle={Conference on Robot Learning},
  year={2022}
}

@inproceedings{janner2022diffuser,
  author={Janner, Michael and Du, Yilun and Tenenbaum, Joshua B. and Levine, Sergey},
  title={Planning with Diffusion for Flexible Behavior Synthesis},
  booktitle={International Conference on Machine Learning},
  year={2022}
}

@inproceedings{ajay2023decisiondiffuser,
  author={Ajay, Anurag and Du, Yilun and Gupta, Abhi and Tenenbaum, Joshua B. and Jaakkola, Tommi and Agrawal, Pulkit},
  title={Is Conditional Generative Modeling All You Need for Decision Making?},
  booktitle={International Conference on Learning Representations},
  year={2023}
}

@inproceedings{wang2023diffusionql,
  author={Wang, Zhendong and Hunt, Jonathan J. and Zhou, Mingyuan},
  title={Diffusion Policies as an Expressive Policy Class for Offline Reinforcement Learning},
  booktitle={International Conference on Learning Representations},
  year={2023}
}

@inproceedings{chi2023diffusionpolicy,
  author={Chi, Cheng and Xu, Zhenjia and Feng, Siyuan and Cousineau, Eric and Du, Yilun and Burchfiel, Benjamin and Tedrake, Russ and Song, Shuran},
  title={Diffusion Policy: Visuomotor Policy Learning via Action Diffusion},
  booktitle={Robotics: Science and Systems},
  year={2023}
}

@inproceedings{psenka2024qsm,
  author={Psenka, Michael and Escontrela, Alejandro and Abbeel, Pieter and Ma, Yi},
  title={Learning a Diffusion Model Policy from Rewards via Q-Score Matching},
  booktitle={International Conference on Machine Learning},
  year={2024}
}

@inproceedings{mao2024diffusiondice,
  author={Mao, Liyuan and Xu, Haoran and Zhan, Xianyuan and Zhang, Weinan and Zhang, Amy},
  title={{Diffusion-DICE}: In-Sample Diffusion Guidance for Offline Reinforcement Learning},
  booktitle={Advances in Neural Information Processing Systems},
  volume={37},
  year={2024},
  doi={10.52202/079017-3136}
}

@inproceedings{ren2024dppo,
  author={Ren, Allen Z. and Lidard, Justin and Ankile, Lars L. and Simeonov, Anthony and Agrawal, Pulkit and Majumdar, Anirudha and Burchfiel, Benjamin and Dai, Hongkai and Simchowitz, Max},
  title={Diffusion Policy Policy Optimization},
  booktitle={International Conference on Learning Representations},
  year={2025}
}

@inproceedings{celik2025dime,
  author={Celik, Onur and Li, Zechu and Blessing, Denis and Li, Ge and Palenicek, Daniel and Peters, Jan and Chalvatzaki, Georgia and Neumann, Gerhard},
  title={{DIME}: Diffusion-Based Maximum Entropy Reinforcement Learning},
  booktitle={Proceedings of the 42nd International Conference on Machine Learning},
  volume={267},
  series={Proceedings of Machine Learning Research},
  pages={6958--6977},
  publisher={PMLR},
  year={2025}
}

@inproceedings{lv2025flowrl,
  author={Lv, Lei and Li, Yunfei and Luo, Yu and Sun, Fuchun and Kong, Tao and Xu, Jiafeng and Ma, Xiao},
  title={Flow-Based Policy for Online Reinforcement Learning},
  booktitle={Advances in Neural Information Processing Systems},
  year={2025}
}

@inproceedings{zhang2025sacflow,
  author={Zhang, Yixian and Yu, Shu'ang and Zhang, Tonghe and Guang, Mo and Hui, Haojia and Long, Kaiwen and Wang, Yu and Yu, Chao and Ding, Wenbo},
  title={{SAC Flow}: Sample-Efficient Reinforcement Learning of Flow-Based Policies via Velocity-Reparameterized Sequential Modeling},
  booktitle={International Conference on Learning Representations},
  year={2026},
  url={https://openreview.net/forum?id=zZvWj4JrYj}
}

@inproceedings{zeng2021transporter,
  author={Zeng, Andy and Florence, Pete and Tompson, Jonathan and Welker, Stefan and Chien, Jonathan and Attarian, Maria and Armstrong, Travis and Krasin, Ivan and Duong, Dan and Sindhwani, Vikas and Lee, Johnny},
  title={Transporter Networks: Rearranging the Visual World for Robotic Manipulation},
  booktitle={Conference on Robot Learning},
  year={2021}
}

@inproceedings{zhao2023aloha,
  author={Zhao, Tony Z. and Kumar, Vikash and Levine, Sergey and Finn, Chelsea},
  title={Learning Fine-Grained Bimanual Manipulation with Low-Cost Hardware},
  booktitle={Robotics: Science and Systems},
  year={2023},
  doi={10.15607/RSS.2023.XIX.016}
}

@inproceedings{brohan2023rt1,
  author={Brohan, Anthony and Brown, Noah and Carbajal, Justice and Chebotar, Yevgen and Chen, Xi and Choromanski, Krzysztof and Ding, Tianli and Driess, Danny and Dubey, Avinava and Finn, Chelsea and others},
  title={{RT-1}: Robotics Transformer for Real-World Control at Scale},
  booktitle={Robotics: Science and Systems},
  year={2023}
}

@inproceedings{chi2024umi,
  author={Chi, Cheng and Xu, Zhenjia and Pan, Chuer and Cousineau, Eric and Burchfiel, Benjamin and Feng, Siyuan and Tedrake, Russ and Song, Shuran},
  title={Universal Manipulation Interface: In-The-Wild Robot Teaching Without In-The-Wild Robots},
  booktitle={Robotics: Science and Systems},
  year={2024}
}

\end{document}